\documentclass[10pt,twocolumn,letterpaper]{article}

\usepackage{cvpr}              
\makeatletter
\@namedef{ver@everyshi.sty}{}
\makeatother
\usepackage{tikz}

\usepackage{graphicx}
\usepackage{amsmath}
\usepackage{amssymb}
\usepackage{booktabs}
\usepackage[most]{tcolorbox}
\usepackage{pifont}
\usepackage{multirow}

\usepackage[font={footnotesize}]{caption}

\usepackage{subcaption}
\usepackage{algorithm}
\usepackage{algpseudocode}

\usepackage[pagebackref,breaklinks,colorlinks]{hyperref}

\usepackage[capitalize]{cleveref}
\crefname{section}{Sec.}{Secs.}
\Crefname{section}{Section}{Sections}
\Crefname{table}{Table}{Tables}
\crefname{table}{Tab.}{Tabs.}

\def\cvprPaperID{723} 
\def\confName{CVPR}
\def\confYear{2023}

\begin{document}

\title{Text-Visual Prompting for Efficient 2D Temporal Video Grounding}

\author{%
  Yimeng Zhang$^{1,2}$, Xin Chen$^{2}$, Jinghan Jia$^{1}$, Sijia Liu$^{1}$, Ke Ding$^{2}$
 \\
  $^{1}$Michigan State University, $^{2}$Applied ML, Intel
\\
  \texttt{\{zhan1853, jiajingh, liusiji5\}@msu.edu}, \\
  \texttt{\{xin.chen, ke.ding\}@intel.com}}

\maketitle

    
\begin{abstract}
  In this paper, we study the problem of temporal video grounding~(\textbf{TVG}), which aims to predict the starting/ending time points of moments described by a text sentence within a long untrimmed video.  Benefiting from fine-grained   3D visual features, the TVG techniques have  achieved remarkable progress in recent years. However, the high complexity of 3D convolutional neural networks (CNNs) makes extracting dense 3D visual features time-consuming, which calls for intensive memory and computing resources. Towards efficient TVG, we propose a novel \textbf{t}ext-\textbf{v}isual \textbf{p}rompting (\textbf{TVP}) framework, which incorporates optimized 
  perturbation patterns (that we call `prompts') into 
  both visual inputs and 
  textual features 
  of a TVG model. In sharp contrast to 3D CNNs,  we show that TVP  allows us to effectively co-train vision encoder and language encoder in  a 2D TVG model   and improves the performance of crossmodal feature fusion using only low-complexity sparse 2D visual features.
  Further, we propose a \textbf{T}emporal-\textbf{D}istance \textbf{IoU}~(\textbf{TDIoU}) loss for efficient learning of TVG.
  Experiments on two   benchmark datasets, Charades-STA and ActivityNet Captions datasets, empirically show that the proposed TVP significantly boosts the performance of 2D TVG (\textit{e.g.}, 9.79\%  improvement on Charades-STA  and 30.77\% improvement on ActivityNet Captions) and achieves $5\times$ inference acceleration over TVG  using 3D visual features. Codes are available at \href{https://github.com/intel/TVP}{\texttt{Open.Intel}}. 

\end{abstract}

\section{Introduction}
\label{sec:intro}

\begin{figure}[!t]
\centerline{\includegraphics[width=1.0\columnwidth]{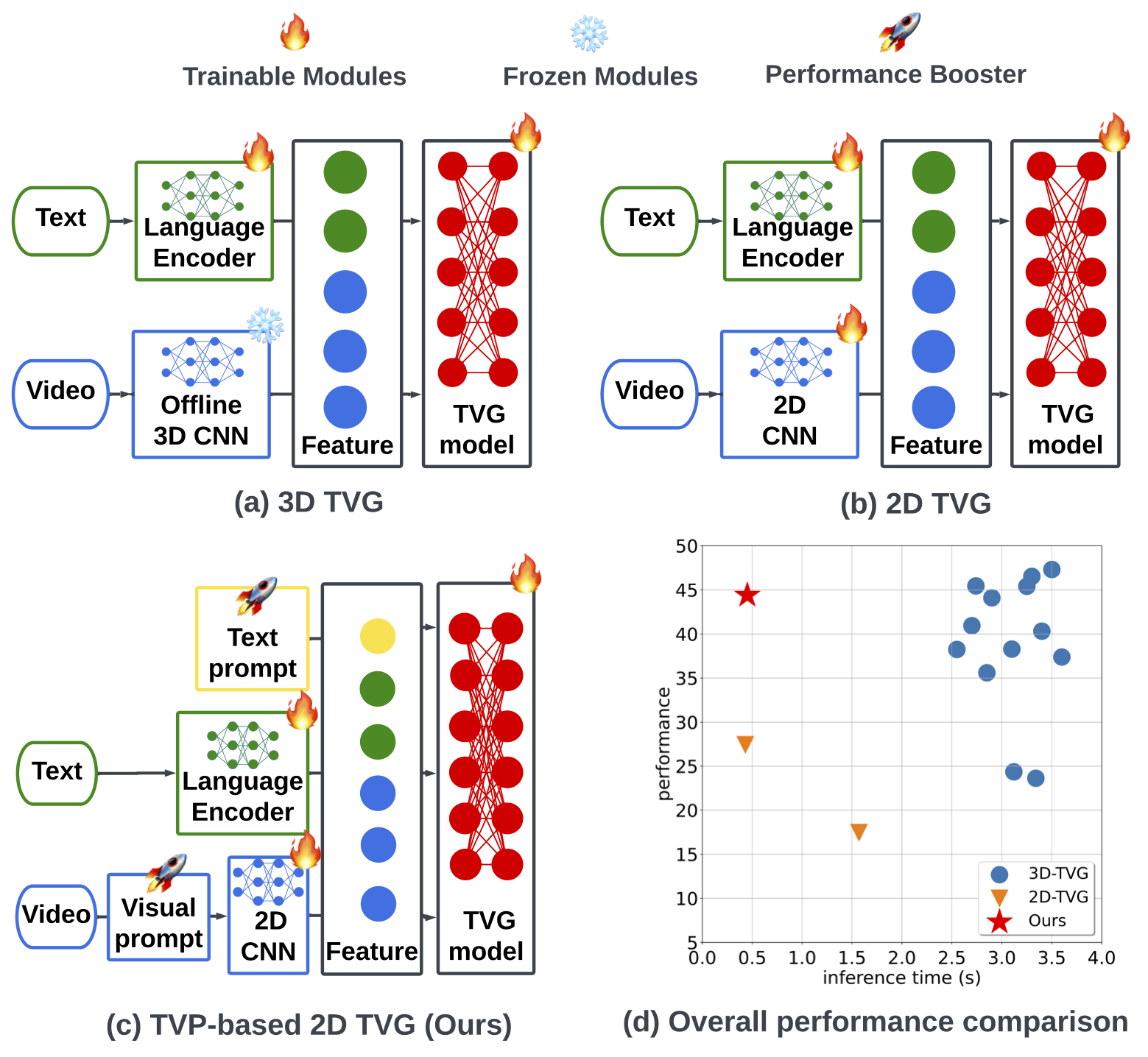}}
\caption{The architecture and performance comparison among TVG methods: \textbf{a)} 3D TVG methods~\cite{gao2017tall,yuan2019find,xiao2021boundary,xiao2021natural, xu2019multilevel, zhang2019man,zeng2020dense,li2021proposal,lu2019debug, ghosh2019excl,zhang2020span,wu2020tree,hahn2019tripping},
\textbf{b)} 2D TVG methods~\cite{ anne2017localizing, chen2019semantic}, and \textbf{c)} TVP-based 2D TVG~(Ours), 
\textbf{d)} overall performance comparison.
Ours is the most efficient~(least inference time) and achieves competitive performance compared to 3D TVG methods.
In contrast to existing TVG methods, which utilize dense video features extracted by non-trainable \textit{offline 3D CNNs} and textual features, our proposed framework utilizes a trainable \textit{2D CNN} as the vision encoder to extract features from sparsely-sampled video frames with a universal set of frame-aware visual prompts and adds text prompts in textual feature space for end-to-end regression-based   modeling.}
\vspace{-5mm}
\label{fig:difference}
\end{figure}

In recent years, we have witnessed great progress on temporal video grounding~(\textbf{TVG})~\cite{lan2021survey,zhang2022elements}. One key to this success comes from the fine-grained dense 3D visual features extracted by 3D convolutional neural networks (CNNs) (\textit{e.g.}, C3D~\cite{tran2015learning} and I3D~\cite{carreira2017quo}) since TVG tasks demand spatial-temporal context to locate the temporal interval of the moments described by the text query. However, due to the high cost of the dense 3D feature extraction, most existing TVG models only take these 3D visual features extracted by offline 3D CNNs as inputs instead of co-training during TVG model training.

Although   models using 3D visual features~({that we call `\textbf{3D methods or models}'}) outperform these using the 2D features~(that we call `\textbf{2D methods or models}'), a unique advantage of 2D methods is that extracting 2D visual features can significantly reduce the cost in TVG tasks
~\cite{gao2017tall,zhang2022elements,lan2021survey,xiao2021boundary,xiao2021natural,zhang2020learning,zeng2020dense,li2021proposal,gao2021evoquer}.
An efficient and lightweight solution with reasonable performance is also demanded in computer vision, NLP, and video-language tasks \cite{zhang2022advancing, zhang2020video, liu2020dynamic, jia2022robustness, zhang2023data,han2020tensor, zhang2022robustify, zafrir2021prune, li2022pruning,zhang2022revisiting}. 
As discussed above, the methods employing 3D video features are challenging to be employed in practical applications. 
It thus has significant practical and economic value to   develop compact 2D solutions for   TVG tasks. In this paper, we   ask:
\begin{tcolorbox}
\textit{How to advance 2D TVG methods so as to achieve comparable results to 3D TVG methods?}
\end{tcolorbox}

To address this problem, we propose a novel \textbf{t}ext-\textbf{v}isual \textbf{p}rompting~(\textbf{TVP}) framework for training TVG models using 2D visual features. As shown in \textbf{Fig.~\ref{fig:difference}}, for existing 2D TVG and 3D TVG methods, they all utilize offline pretrained vision encoders and language encoders to perform feature extraction. In contrast, our proposed TVP framework is end-to-end trainable. Furthermore, benefiting from text-visual prompting and cross-modal pretraining on large-scale image-text datasets, our proposed framework could achieve comparable performance to 3D TVG methods with significant inference time acceleration.

\begin{figure}[t]
    \centering
    \begin{subfigure}[t]{0.15\textwidth}
           \centering
           \includegraphics[width=\textwidth]{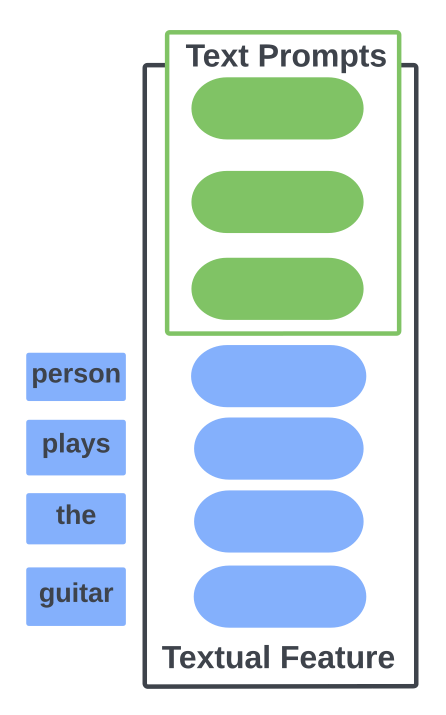}
            \caption{Text Prompts}
            \label{fig:5a}
    \end{subfigure}
    \begin{subfigure}[t]{0.27\textwidth}
            \centering
            \includegraphics[width=\textwidth]{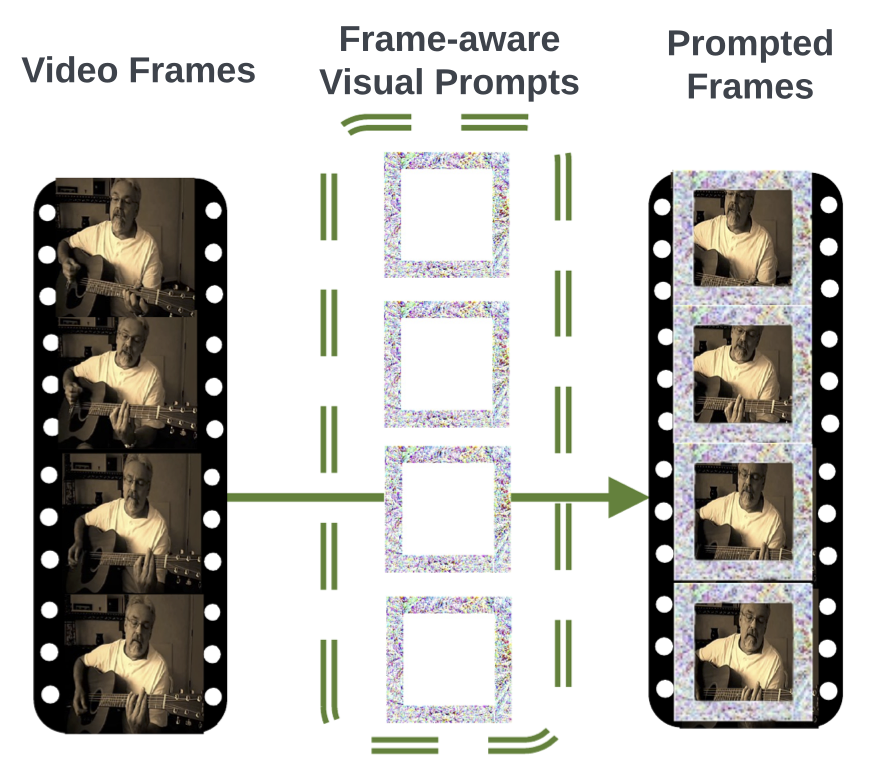}
            \caption{Frame-aware Visual Prompts}
            \label{fig:5b}
    \end{subfigure}
    \caption{Text-visual prompting illustration. (a) Text prompts are directly applied in the feature space. (b) A set of   visual prompts are applied to  video frames in order.
    }
    \vspace*{-2mm}
\label{fig:tvp_vis}
\end{figure}

Conventionally,
TVG methods consist of three stages: 
\ding{172} extracting feature from visual and text inputs; 
\ding{173} multi-modal feature fusion; 
\ding{174} cross-modal modelling. 
In contrast to conventional methods, TVP incorporates  optimized input perturbation patterns  ({that we call `\textbf{prompts}'}) into both visual inputs and textual features of a TVG model.
We apply trainable parameters in the textual features as text prompts and develop a universal set of frame-aware patterns as visual prompts.
Specially, we sample a fixed number of frames from a video and optimize text prompts for the input query sentence and a set of visual prompts for frames with different temporal locations during training.
During testing, the same set of optimized visual prompts and textual prompts are applied to all test-time videos.
We refer readers to \textbf{Fig.~\ref{fig:tvp_vis}} for illustrations of visual prompts and text prompts introduced.
To the best  of our knowledge, our work makes the first attempt to utilize prompt learning to successfully improve the performance of regression-based TVG tasks using 2D visual features.

Compared to 3D CNNs, 2D CNNs loses spatiotemporal information of the video during feature extraction. 
Inspired by the success of transformers on the vision-language tasks~\cite{li2019visualbert,huang2020pixel, qi2020imagebert,lu2019vilbert, su2019vl, tan2019lxmert,chen2020uniter} and the recent application of prompt learning  to transformers in both vision and language domains~\cite{jiang2020can,li2021prefix,lester2021power,liu2021gpt,bahng2022exploring,khattak2022maple},
we choose   transformer as our base TVG model and propose to utilize prompts to compensate for the lack of spatiotemporal information in 2D visual features.
Furthermore, we develop a \textbf{T}emporal-\textbf{D}istance IoU (\textbf{TDIoU}) loss for training our proposed framework.
There are two aspects that distinguish our proposed framework from existing works. \underline{First}, our proposed framework is designed to boost the performance of the regression-based TVG methods utilizing 2D CNNs as the vision encoder, not for transfer learning~\cite{houlsby2019parameter, ju2021prompting, bahng2022exploring} 
\underline{Second}, our proposed framework utilizes 2D CNN to extract visual features from sparsely-sampled video frames, which requires less memory and is easier to be applied in practical applications compared to 3D methods
~\cite{xiao2021boundary,xiao2021natural,zhang2020learning,wu2020tree,zeng2020dense,li2021proposal}, especially for long videos. Furthermore, thanks to the compact 2D CNN as the vision encoder, our proposed framework could implement the language encoder and visual encoder co-training for better multimodal feature fusion.
In summary, the   \textbf{contributions} of this work are unfolded below: 
\begin{itemize}
\item We propose an effective and efficient framework to train 2D TVG models, in which we leverage TVP (text-visual prompting) to improve the utility of sparse 2D visual features without resorting to costly 3D features. To the best of our knowledge, it is the first  work to expand the application of prompt learning for resolving TVG problems. Our method   outperforms all of 2D methods and achieves competitive performance to 3D TVG methods.

\item Technology-wise, we integrate visual prompt with text prompt
to co-improve the effectiveness of 2D visual features. On top of that, we propose
TDIoU (temporal-distance IoU)-based prompt-model co-training method to obtain high-accuracy 2D TVG models.

\item 
Experiment-wise, we show the empirical success of our proposal to 
boost the performance of 2D TVG on Charades-STA and ActivityNet Captions datasets, \textit{e.g.}, 9.79\% improvement in Charades-STA, and  $30.77\%$ in ActivityNet-Captions together with ${5 \times}$ inference time acceleration over 3D TVG methods. 

\end{itemize}

\section{Related Work}
\label{sec:related}

\noindent
\textbf{Video Temporal Grounding (TVG).}~The objective of the TVG is to predict the starting/ending time points of target moments within an untrimmed video, which is described by a text sentence.  
Early TVG solutions~\cite{gao2017tall, hendricks2018localizing,liu2018cross, zeng2021multi,xu2019multilevel, chen2019semantic, xiao2021boundary} mainly employ two-stage ``propose-and-rank" pipeline: \ding{172} Propose: utilize sliding windows or proposal network to generate proposal candidates from the input video. \ding{173} Rank: the proposed candidates would be ranked according to the text query, and then the proposal with the highest ranking would be the final prediction decision.
In contrast to proposal-based methods, regression-based methods~\cite{yuan2019find, ghosh2019excl,zeng2020dense} directly predict the starting/ending time points of the target moments without ranking massive proposal candidates. Thus, regression-based methods are much faster than proposal-based methods, which is one reason why our work focuses on the regression-based TVG.
Furthermore, reinforcement learning (RL)-based methods formulate the TVG task as a sequence of decisions to make ~\cite{hahn2019tripping,wu2020tree}. In particular, they train an agent to control the movement of a window by shifting or scaling. During training, the agent would be rewarded or punished based on whether the window is close to the target moment after an adjustment.

\noindent
\textbf{Temporal Action Detection (TAD).} 
TAD aims to determine whether predefined actions occur in a video and to predict the corresponding time intervals during which these actions occur~\cite{simonyan2014two,xie2018rethinking,tran2015learning,qiu2017learning,wang2018non,feichtenhofer2020x3d,feichtenhofer2019slowfast}. Different from TVG, the input of TAD is only a video. In other words, TAD only requires a semantic understanding of videos. Compared to TAD, TVG is more challenging since it requires a semantic understanding of both videos and natural languages. Furthermore, TVG needs to process the multimodal interaction between videos and natural languages.

\noindent
\textbf{Text Prompting.} Prompting has recently achieved great success in the  domain   of natural language processing \cite{petroni2019language,schick2020exploiting,schick2020few,schick2020s,jiang2020can,shin2020autoprompt,li2021prefix,lester2021power,wallace2019universal,liu2021gpt}. Text prompting is a process that leverages a data-agnostic  perturbation operation 
applied  to   text inputs or their embeddings  to improve the performance of the downstream task. The simplest way is to construct an input context template originating from human contemplation   \cite{petroni2019language,schick2020exploiting,schick2020few,schick2020s}. Although the manually-crafted context templates are simple and interpretable, they are typically not the optimal input prompts. To tackle this issue, other work has
focused on searching the optimal prompting in the discrete input space \cite{jiang2020can,shin2020autoprompt,wallace2019universal} or in the language model's embedding space      \cite{li2021prefix,lester2021power,liu2021gpt}.

\noindent
\textbf{Visual Prompting.} Inspired by the idea of prompt learning in NLP  \cite{li2021prefix}, visual prompting~(VP) was first proposed by Bahng \textit{et. al.}~\cite{bahng2022exploring} to reprogram     a source vision model (\textit{e.g.}, ImageNet-pretrained classifier) to accomplish downstream target tasks (\textit{e.g.}, CIFAR-10 image classification). VP shares almost the same idea with the    model reprogramming technology in the vision domain \cite{chen2022model,elsayed2018adversarial,tsai2020transfer,zheng2021adversarial,yang2021voice2series, zhang2022fairness, chen2022visual, chen2022understanding}, which  incorporates a universal input perturbation  into testing data so as to improve a desired performance metric, \textit{e.g.}, target task accuracy, robustness, and fairness.

\noindent
\textbf{Multi-Modal Prompting.}
Although visual prompting and text prompting have recently attracted much attention,  they are under-explored in the multi-modal learning, especially on the temporal video grounding task.  The existing works \cite{bahng2022exploring,khattak2022maple,yao2021cpt} mainly focus on integrating text   and visual prompts with the CLIP (Contrastive Language–Image Pretrained) model 
to improve downstream tasks with imagery data.  The problem of 
  multi-modal prompting  in the video understanding task has not been studied. In this paper, we for the first time develop the text-visual prompting technique  to   improve the performance of   temporal video grounding using 2D  visual features.

\section{Methods} \label{sec:method}

In this section, we begin with the problem formulation of regression-based TVG. Then we demonstrate the design of TVP (text-visual prompts) and present the overview  of our proposed TVP framework.

\subsection{Problem Definition} 
\label{sec:problem}

Let  
$\mathbf{v}
\in \mathbb R^{N_\mathrm{vid} \times C \times H \times W}$  
be an untrimmed \underline{vid}eo consisting of a sequence of $N_\mathrm{vid}$ video frames, 
and 
$\mathbf{s} \in \mathbb R^{N_\mathrm{tex}}$  
be a \underline{tex}t query consisting of a sequence of $N_\mathrm{tex}$ language tokens. Here,  the video-query pair $(\mathbf{v}, \mathbf{s} )$ belongs to a video-language dataset $\mathcal D$.
Given   $\mathbf{v}$ and   $\mathbf{s}$, TVG aims to {pre}dict the time interval 
${\mathbf{\hat{T}}}= (\hat{t}_\mathrm{sta},\hat{t}_\mathrm{end})$ of the target video moments described by the query $\mathbf{s}$. 
The TVG model that fuses the vision-language modalities can be described as:
\vspace*{-1mm}
\begin{align}
\mathbf{\hat{T}}=  f(~ g_\mathrm{tex}(\mathbf{s}), ~ g_\mathrm{vid}(\mathbf{v})~),
\label{eq: defintion}
\vspace*{-3mm}
\end{align}
where $f$ denotes TVG model, and $g_\mathrm{vid}$ and $g_\mathrm{tex}$ represent vision encoder and language encoder, respectively. 

\begin{figure*}[htb]
\centerline{\includegraphics[width=2.0\columnwidth]{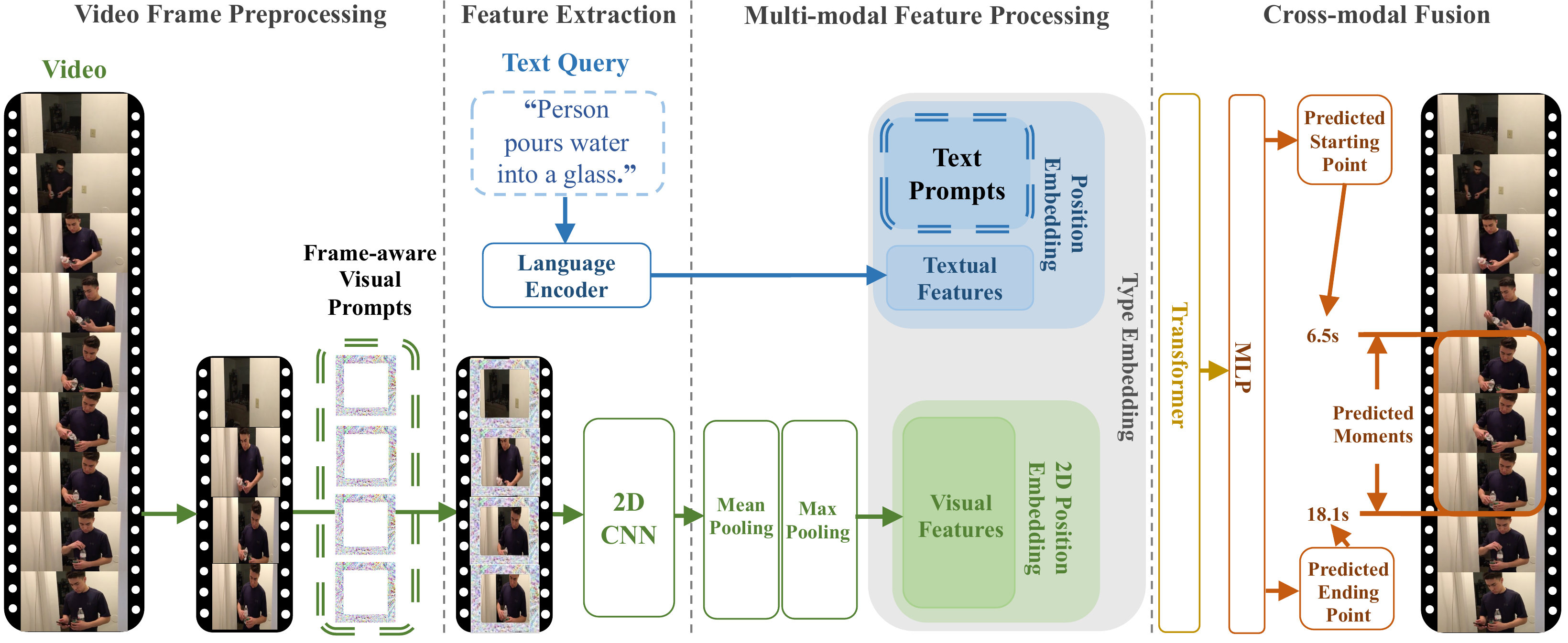}}
\caption{Overview of our proposed TVP (text-visual prompting) framework for 2D TVG (temporal video grounding). The whole process contains four phases: \ding{182}~Video frame preprocessing: uniformly sample frames from input video and apply a set of frame-aware visual prompts to the sampled frames in order; 
\ding{183}~Feature extraction: 2D CNN extracts features from sampled video frames with visual prompts, and the language encoder extracts textual features. In addition, the visual features would be spatially downsampled and temporally fused by max pooling and mean pooling, respectively.  
\ding{184}~Multimodal feature processing: after spatial downsampling and temporal fusion, the 2D visual features would be integrated into the prompted textual features. 
\ding{185}~Crossmodal fusion: the multimodal features would be processed by a 12-layer transformer encoder, and MLP would predict the starting/ending time points of the target moment.}

\label{fig:overall}
\end{figure*}

\subsection{\textbf{TDIoU} Loss Function} 
Conventionally, the TVG model can be learned by minimizing the \textbf{temporal IoU loss} $\mathcal{L}_ \mathrm{tIoU}$ defined   below:
\begin{align}
\mathcal{L}_\mathrm{tIoU}  
=   \left( 1 - \frac{
\mathbf{\hat{T}}(\boldsymbol \theta) \bigcap \mathbf{T}}{{\mathbf{\hat{T}}(\boldsymbol \theta)} \bigcup \mathbf{T}} \right),
\label{eq: problem}
\vspace*{-3mm}
\end{align}
where 
for ease of notation
let $\boldsymbol{\theta}$ denote all the trainable   parameters involved in \eqref{eq: defintion}, and $\mathbf{T} = (t_\mathrm{sta},t_\mathrm{end})$ is the label (\textit{i.e.}, the ground-truth time interval) of the target moment associated with the  input video-query pair   $(\mathbf{v}, \mathbf{s} )$. 
The rationale behind \eqref{eq: problem} is to 
maximize the overlapping between the predicted time interval and its ground truth.

However, for non-overlapping cases, the temporal IoU loss $\mathcal{L}_ \mathrm{tIoU}$ would encounter a gradient vanishing problem. 
Inspired by \cite{zheng2020distance}, we develop a novel \textbf{TDIoU} (Temporal-Distance IoU)
loss for training our proposed TVG models by incorporating the normalized central time point distance and duration difference between the predicted video clips and the target video clips. We elaborate on the proposed loss below.

\noindent
\textbf{\underline{Dis}tance Loss $\mathcal{L}_\mathrm{dis}$.}
To avoid the gradient vanishing problem caused by the non-overlapping case, we involve distance loss $\mathcal{L}_\mathrm{dis}$ to directly minimize the normalized central time point distance. In addition, we add a threshold $\alpha_1$ to prevent oscillation in the later training phase. The distance loss is then given by:

\vspace*{-2mm}
{\small
\begin{align}
\mathcal{L}_\mathrm{dis} = \max \left( \frac{
\vert\left( t_\mathrm{sta} + t_\mathrm{end} \right)/2 -
\left( \hat{t}_\mathrm{sta} + \hat{t}_\mathrm{end} \right)/2\vert
}
{\vert \mathbf{\hat{T}} \bigcup  \mathbf{T} \vert}, ~\alpha_1 \right),
\label{eq: dloss}
\vspace*{-3mm}
\end{align}}%
where recall that $\mathbf{T} = (t_\mathrm{sta},t_\mathrm{end})$,   $\mathbf{\hat{T}}$ is predicted by the TVG model \eqref{eq: defintion}, and we choose $\alpha_1 = 0.2$ in experiments.

\noindent
\textbf{\underline{Dur}ation Loss $\mathcal{L}_\mathrm{dur}$.}  
The introduction of distance loss $\mathcal{L}_\mathrm{dis}$ avoids the gradient vanishing problem but only considers the central time point distance.  Yet, this may not be precise enough. For example, even if   the central time points are completely overlapped, the duration of two video clips may not be identical. 
Inspired by the above, we propose the duration loss:
\vspace*{-1mm}
\begin{align}
\mathcal{L}_\mathrm{dur} = 
\max \left( \frac{
\vert \mathbf{T} - \mathbf{\hat{T}(\boldsymbol \theta)} \vert
}
{\vert \mathbf{T} \vert}, ~\alpha_2 \right),
\label{eq: durloss}
\vspace*{-3mm}
\end{align}%
where 
$\alpha_2$ is the precision tolerance threshold and set by 0.4 in our experiments.

Finally, the proposed Temporal-Distance
IoU (TDIoU) loss is given by
\vspace*{-1mm}
\begin{align}
\mathcal{L} 
=  \mathcal{L}_{\mathrm{tIoU}} + \beta_1 \mathcal{L}_{\mathrm{dis}} + \beta_2 \mathcal{L}_{\mathrm{dur}},
\label{eq: dur_los}
\end{align}%
where 
$\beta_1>0 $ and $\beta_2 >0 $ are regularization parameters.

\subsection{Text-Visual Prompt Design}
Inspired by the application of prompts on transformers~\cite{houlsby2019parameter, li2021prefix, bahng2022exploring,li2022bridge}, we propose jointly text-visual prompting to boost the performance of our models, 
in which prompts are optimized perturbation patterns.
To improve data processing efficiency, 
we uniformly \underline{sam}ple video frames from the untrimmed video $\mathbf v$ to obtain $\mathbf{v}_\mathrm{sam} \in \mathbb R^{N_\mathrm{sam} \times C \times H \times W}$, where  $N_\mathrm{sam}$ is the number of sampled video frames.
In addition, we introduce a set of frame-aware \underline{v}isual \underline{p}rompts
$\boldsymbol \delta_\mathrm{vp} \in \mathbb{R}^{N_\mathrm{sam} \times d_\mathrm{vp}}$   in the pixel space of sampled video frames $\mathbf{v}_\mathrm{sam}$, and introduce \underline{t}ext \underline{p}rompts $\boldsymbol \delta_\mathrm{tp} \in \mathbb{R}^{N_\mathrm{tp} \times d_\mathrm{tp}}$   in the textual feature space.
By incorporating video frame sampling and text-visual prompts into
 the TVG model \eqref{eq: defintion}, we  obtain:
\begin{align}
(\hat{t}_\mathrm{sta},\hat{t}_\mathrm{end})
= f( ~
\boldsymbol \delta_\mathrm{tp},~
g_{\mathrm{tex}}(\mathbf{s}), ~ g_{\mathrm{vid}}(\mathbf{v}_\mathrm{sam} + \boldsymbol \delta_\mathrm{vp}) 
~).
\vspace*{-3mm}
\label{eq: prompt_defin}
\end{align}
\vspace*{-3mm}

Given a  pre-trained 2D TVG model $f$, the objective of text-visual prompting (TVP) is to learn 
a universal set of
visual prompts $\boldsymbol \delta_\mathrm{vp}$ and   text prompts $\boldsymbol \delta_\mathrm{tp}$ to be integrated into sampled video frames and textual features, respectively. 
Specially, a set of different visual prompts 
are applied to uniformly-sampled frames of one untrimmed video in order. During training, only the set of visual prompts and text prompts are updated through backpropagation.
During finetuning, prompts are frozen, and the parameters of the TVG model and encoders are updated. During testing, the set of optimized visual prompts and the optimized text prompts are applied to all test-time video-query pairs.

\subsection{Framework}
\label{sec:pipeline}
Inspired by the success of transformers in vision-language tasks, we choose   ClipBERT~\cite{lei2021less} as the base model for 2D TVG.
Extended from ClipBERT,  the input of our  regression-based TVG model would be describable sentences and uniformly sampled frames of one untrimmed video as shown in \textbf{Fig.~\ref{fig:overall}}.
Then, the predicted starting and ending time points of the target video clip would be model outputs. 
As described in \textbf{Algorithm~\ref{alg:tvp}}, there are four phases of our proposed TVP framework: 
\ding{182} \textbf{\underline{Video frame preprocessing}}: We obtain sparsely-{sam}pled frames $\mathbf{v}_\mathrm{sam}$
from one input untrimmed video $\mathbf{v}$, and apply universal frame-aware visual prompts $\boldsymbol{\delta}_{\mathrm{vp}}$ on top of frames at the padding location. 
\ding{183} \textbf{\underline{Feature extraction}}: 
2D vision 
encoder (first 5 ConvBlock of ResNet-50) $g_\mathrm{vid}$ 
and language encoder (a trainable word embedding layer) $g_\mathrm{tex}$ would extract features from the prompted frames $\mathbf{v}^{\prime}_\mathrm{sam} $
and textual inputs $\mathbf{s}$, respectively. 
\ding{184} \textbf{\underline{Multimodal feature processing}}:
Following the setting of Pixel-BERT~\cite{huang2020pixel}, the 2D visual features $\mathbf{Q}_\mathrm{vid}$
are downsampled spatially by a $2\times2$ max-pooling layer and fused temporally by a mean-pooling layer. 
Then, text prompts $\boldsymbol{\delta}_{\mathrm{tp}}$ are integrated into textual features $\mathbf{Q}_\mathrm{tex} $.
 In addition, trainable 2D visual position embeddings 
$\mathbf{M}_\mathrm{2D} $
and textual position embeddings 
$\mathbf{M}_\mathrm{pos}$
are applied to the processed 2D visual features $\mathbf{Q}^{\prime}_{\mathrm{vid}} $
and prompted textual features $\mathbf{Q}^{\prime}_\mathrm{tex}$, respectively \cite{lei2021less, devlin2018bert}. 
Afterwards, the processed and position-encoded 2D visual features  $\mathbf{Q}^{\prime\prime}_{\mathrm{vid}}$ are flattened and integrated into prompted and position-encoded textual features $\mathbf{Q}^{\prime\prime}_\mathrm{tex}$. Moreover, type embeddings $\mathbf{M}_\mathrm{type}$
would be added to the integrated multimodal features $\mathbf{Q}_{\mathrm{all}}$ 
to indicate the source type of features.
\ding{185} \textbf{\underline{Crossmodal fusion}}: A 12-layer transformer~\cite{devlin2018bert} is utilized for crossmodal fusion on  $\mathbf{Q}_{\mathrm{all}}$, and then multilayer perceptron (MLP) ending with sigmoid function is used as the prediction head 
to process the last-layer \underline{c}ross\underline{m}odal representation $\mathbf{Q}_\mathrm{CM}$ 
of the transformer for generating the predicted starting/ending time points  $ (\hat{t}_\mathrm{sta}, \hat{t}_\mathrm{sta})$ of the target moments described by the text query input.

\begin{algorithm}[htb]
\caption{Overview of TVP  framework} 
\label{alg:tvp}
\begin{algorithmic}[1]
\Require 
vision encoder $g_\mathrm{vid}$, 
language encoder $g_\mathrm{tex}$, 
position embeddings $\mathbf{M}_\mathrm{pos}$,
2D position embeddings $\mathbf{M}_\mathrm{2D}$,
type embeddings $\mathbf{M}_\mathrm{type}$,
transformer $f$, 
prediction head $MLP$,
visual prompts $\boldsymbol{\delta}_{\mathrm{vp}}$, 
text prompts $\boldsymbol{\delta}_{\mathrm{tp}}$
\Ensure 
Predicted time interval  $\mathbf{\hat{T}} = (\hat{t}_\mathrm{sta}, \hat{t}_\mathrm{end})$
\Statex \qquad \textbf{Phase \ding{182}: Video frame preprocessing}
\State $\mathbf{v}_\mathrm{sam} \gets$  uniformly sample video frames from an untrimmed video $\mathbf{v}$
\State $\mathbf{v}^{\prime}_\mathrm{sam} \gets$ apply visual prompts $\boldsymbol{\delta}_{\mathrm{vp}}$ to the sampled video frames $\mathbf{v}_\mathrm{sam}$
\Statex \qquad \textbf{Phase \ding{183}: Feature Extraction}
\State $ \mathbf{Q}_\mathrm{vid} =  g_\mathrm{vid}(\mathbf{v}^{\prime}_\mathrm{sam}) \gets $ extracting 2D visual features 
\State $ \mathbf{Q}_\mathrm{tex} =  g_\mathrm{tex}(\mathbf{s}) \gets $ extracting textual features 
\Statex \qquad \textbf{Phase \ding{184}: Multimodal feature processing}
\State $ \mathbf{Q}^{\prime}_\mathrm{vid} \gets $ apply spatial downsampling and temporal fusion to 2D visual features $ \mathbf{Q}_\mathrm{vid}$
\State $ \mathbf{Q}^{\prime}_\mathrm{tex} \gets $ apply text prompts $\boldsymbol{\delta}_{\mathrm{tp}}$ to textual features $\mathbf{Q}_\mathrm{tex}$
\State $ \mathbf{Q}^{\prime\prime}_\mathrm{vid} \gets$ add 2D visual position embeddings $\mathbf{M}_\mathrm{2D}$ on the processed 2D visual features $\mathbf{Q}^{\prime}_\mathrm{vid}$
\State $ \mathbf{Q}^{\prime\prime}_\mathrm{tex} \gets $ add position embeddings $\mathbf{M}_\mathrm{pos}$ to prompted textual features $\mathbf{Q}^{\prime}_\mathrm{tex}$
\State $ \mathbf{Q}_\mathrm{all} \gets $ integrate the processed and position-encoded textual features $\mathbf{Q}^{\prime\prime}_\mathrm{tex}$ and the processed and position-encoded 2D visual features $ \mathbf{Q}^{\prime\prime}_\mathrm{vid}$
\State  $ \mathbf{Q}_\mathrm{all} + \mathbf{M}_\mathrm{type} \gets $  add type embeddings $\mathbf{M}_\mathrm{type}$ to the integrated multimodal features $\mathbf{Q}_\mathrm{all}$
\Statex \qquad \textbf{Phase \ding{185}: Crossmodal fusion}
\State   $ \mathbf{Q}_\mathrm{CM} = f(\mathbf{Q}_\mathrm{all} + \mathbf{M}_\mathrm{type}) \gets $ implement \underline{c}ross\underline{m}odal fusion through transformer $f$
\State $ (\hat{t}_\mathrm{sta}, \hat{t}_\mathrm{end})=MLP(\mathbf{Q}_\mathrm{CM})\gets $ prediction head generates the predicted time interval according to crossmodal representation $ \mathbf{Q}_\mathrm{CM}$
\end{algorithmic}
\end{algorithm}

\section{Experiments}
\label{sec:exp}
In this section, we demonstrate the effectiveness of our proposed TVP 
framework
on Charades-STA and ActivityNet Captions datasets.

\begin{table}[t]
\caption{
Statistics of TVG benchmark datasets (Charades-STA and ActivityNet Captions datasets).
}
\vspace{-4mm}
\center
\resizebox{0.47\textwidth}{!}{
\begin{tabular}{c|c|c} 
\toprule
\midrule 
 Dataset & Charades-STA & ActivityNet Captions\\
\midrule 
 Domain & Indoor Activity &
 Indoor/Outdoor Activity \\
\midrule 
 \# Videos & $6,672$ & $14,926$ \\
 Avg. Video Length ($second$) & $30.6$ & $117.6$\\
\midrule 
 \# Moments & $11,767$ & $71,953$ \\
 Avg. Moment Length ($second$) & $8.1$ & $37.1$ \\
\midrule 
 Vocabulary Size  & $1,303$ & $15,505$\\
 \# Queries & $16,124$ & $71,953$ \\
 Avg. Query Length ($word$) & $7.2$ & $14.4$ \\
\midrule 
\bottomrule
\end{tabular}}
\label{tab:dataset}
\end{table}

\subsection{Experiment Setup}
\noindent
\textbf{Datasets.}  ~ 
The evaluations are implemented on two standard benchmark datasets for TVG task, Charades-STA~\cite{gao2017tall} and ActivityNet Captions~\cite{krishna2017dense}. \textbf{Tab.~\ref{tab:dataset}} summarizes the details of both datasets. \textbf{Charades-STA} dataset
contains $6,672$ videos and $16,124$ text queries in total. 
The average length of videos is $30.6s$, and the average length of text query is $7.2~words$ . The average length of moments corresponding to the text query is $8.1s$. Following the same dataset split as \cite{gao2017tall} for fair comparisons, there are $12,408$ video-query pairs for training and $3,720$ pairs for testing. \textbf{ActivityNet Captions} dataset contains $14,926$ videos and $71,953$ text queries in total. 
The average length of videos is $117.6s$, and the average length of text query is $14.4~ words$. The average length of moments corresponding to the text query is $37.1s$. ActivityNet Captions dataset is split into training set, validation set, and testing set in a $2:1:1$ ratio. 
Since the testing set is withheld for competition, only a training set and two validation sets (\textit{val1} and \textit{val2}) can be accessed publicly. For fair comparisons, we evaluate our proposed framework on \textit{val1}.

\noindent
\textbf{Baselines.}  We compare our proposal with $15$ baseline methods: 
\ding{172}~\textbf{Proposal-based}: CTRL~\cite{gao2017tall}, MCN~\cite{anne2017localizing}, SAP~\cite{chen2019semantic}, BPNet~\cite{xiao2021boundary}, LPNet~\cite{xiao2021natural}, QSPN~\cite{xu2019multilevel}, MAN~\cite{zhang2019man}; 
\ding{173}~\textbf{Proposal-free}:
ABLR~\cite{yuan2019find},
DRN~\cite{zeng2020dense}, CPNet~\cite{li2021proposal}, DEBUG~\cite{lu2019debug}, ExCL~\cite{ghosh2019excl}, VSLNet~\cite{zhang2020span};
\ding{174}~\textbf{Reinforcement learning}: TSP-PRL~\cite{wu2020tree}, TripNet~\cite{hahn2019tripping}.

\noindent
\textbf{Evaluation metrics.} Following \cite{gao2017tall}, we adopt 
Acc(R@1, IoU=m)
as the performance evaluation metric, which represents the percentage accuracy of top-$1$ predicted moments whose  tIoU (temporal IoU) with the ground-truth moment is larger than $m$.
By convention, we consider  the following tIoU threshold values  $m=\{0.3, 0.5, 0.7\}$. 

\noindent
\textbf{Crossmodal pretraining setup.} 
Our 2D vision encoder (ResNet-50) is initialized with the weight from grid-feat~\cite{jiang2020defense}, which can extract effective grid features from visual inputs. In addition, both the language encoder and 12-layer transformer are initialized with the BERT-base model weight~\cite{devlin2018bert}, which are pretrained on English Wikipedia and BookCorpus~\cite{zhu2015aligning}. 
Thanks to the compact 2D vision encoder, TVP (our proposal) is able to directly utilize image-text pairs for end-to-end training. Since the benefits of cross-modal pretraining has been demonstrated by \cite{huang2020pixel,lu2019vilbert,tan2019lxmert}, our base model is pretrained on two large-scale image-text datasets, which are Visual Genome Captions~\cite{krishna2017visual} and COCO Captions~\cite{chen2015microsoft}. To be more specific, image-text matching~\cite{tan2019lxmert,lu2019vilbert} and masked language modeling~\cite{devlin2018bert} are employed for cross-modal pretraining.

\noindent
\textbf{Implementation setup.} 
For video inputs, we uniformly sample $N_\mathrm{sam}$ frames from a video ($N_\mathrm{sam}=48$ for Charades-STA and $N_\mathrm{sam}=64$  for ActivityNet Captions). In addition, all video frames are resized to have a maximum longer side of $448$ with an original aspect ratio, and then the frames are zero-padded to $448\times448$.  The default visual prompt sizes for both dataset are $96$.  The default text prompt sizes are $10$ and $20$ for Charades-STA and ActivityNet Captions, respectively.  We utilize the first 5 ConvBlocks of ResNet-50 as the 2D vision encoder and a trainable embedding layer as the language encoder for both Charades-STA and ActivityNet Captions datasets.
For text queries, all word tokens are maintained after lower-case conversion and tokenization.
We use AdamW~\cite{loshchilov2017decoupled} for end-to-end model training, with $\beta_1 = 1.0$, $\beta_2=0.1$, $\alpha_1 = 0.2$, $\alpha_2=0.4$. Initial learning rates are  $1e-1$ and $5e-7$ for prompt training and model finetuning, respectively. In addition, the learning rate linearly decays to $0$ with the first $10\%$ training step for warmup. Our experiments are implemented in PyTorch~\cite{paszke2019pytorch}, and models and prompts are finetuned separately for $12$ epochs with the mixed precision on 8 NVIDIA V100 GPUs. 

\subsection{Experiment Results} ~

\begin{table}[t]
\caption{
Performance comparison of different thresholds $m$ on the  Charades-STA dataset. 
}
\vspace*{-4mm}
\center
\resizebox{.47\textwidth}{!}{
\begin{tabular}{c|c|c|ccc} 
\toprule[1pt]
\midrule
 \multirow{2}{*}{Type} & \multirow{2}{*}{Method} & Visual  & & Acc(R@1, IoU=$m$) &  \\
 & & Feature &  $m$=0.3 & $m$=0.5& $m$=0.7  \\
\midrule
\multirow{13}*{3D TVG} & CTRL~\cite{gao2017tall} & C3D & - & 23.63 & 8.89 \\
~&ABLR~\cite{yuan2019find} & C3D & - & 24.36 & 9.01  \\
~& BPNet~\cite{xiao2021boundary} & C3D & 55.46 &38.25 &20.51\\
~ & LPNet~\cite{xiao2021natural} & C3D & 59.14 & 40.94 & 21.13 \\
~ & QSPN~\cite{xu2019multilevel} & C3D & 54.70 & 35.60 & 15.80 \\
~& TSP-PRL~\cite{wu2020tree} & C3D & - &  45.45 & 24.75 \\
~&TripNet~\cite{hahn2019tripping} & C3D & 54.64 & 38.29 & 16.07 \\
~& DRN~\cite{zeng2020dense} & C3D & - & 45.40 & 26.40 \\
~& CPNet~\cite{li2021proposal} & C3D & - & 40.32 & 22.47 \\
~& DEBUG~\cite{lu2019debug} & C3D & 54.95 & 37.39 & 17.92 \\
~& ExCL~\cite{ghosh2019excl} & I3D & 61.50 & 44.1 & 22.40 \\
~& VSLNet~\cite{zhang2020span} & I3D & 64.30 &  \textbf{47.31} & \textbf{30.19} \\
~& MAN~\cite{zhang2019man}& I3D & - & 46.53 & 22.72 \\
\midrule
\multirow{2}*{2D TVG} & MCN~\cite{anne2017localizing} & VGG & - & 17.46 & 8.01  \\
~& SAP~\cite{chen2019semantic} & VGG & - & 27.42 & 13.36  \\ 
\midrule
\multicolumn{6}{c}{\textbf{Ours}} \\
\midrule
\multirow{4}{*}{\begin{tabular}[c]{@{}c@{}} TVP-Based\\ 2D TVG\end{tabular}} 

&Base w/o prompts & \multirow{4}{*}{ResNet} & 61.29 & 40.43 &  19.89  \\
~&Base + Visual Prompts &  & 65.38 & 44.31 & 20.22 \\
~&Base + Text Prompts & & 65.81 &  43.44 & 20.65  \\
~&Base + Both Prompts & & $\textbf{65.92}$ & 44.39& 21.51  \\
\midrule
\bottomrule[1pt]
\end{tabular}}
\label{tab:charades_result}
\end{table}

\vspace*{2mm}

\begin{table}[t]
\caption{
Performance comparison of different thresholds $m$ on the ActivityNet Captions dataset. 
}
\vspace*{-4mm}
\center
\resizebox{.47\textwidth}{!}{
\begin{tabular}{c|c|c|ccc} 
\toprule[1pt]
\midrule
 \multirow{2}{*}{Type} & \multirow{2}{*}{Method} & Visual  & & Acc(R@1, IoU=$m$) &  \\
  & & Feature & 
  $m$=0.3 & $m$=0.5& $m$=0.7  \\
  \midrule
  \multirow{12}{*}{3D TVG}& CTRL~\cite{gao2017tall} & C3D & 28.70 & 14.00 & - \\
  & BPNet~\cite{xiao2021boundary} & C3D & 59.98 & 42.07 & 24.69 \\
  & LPNet~\cite{xiao2021natural} & C3D &  \textbf{64.29} & \textbf{45.92} & 25.39\\
  & QSPN~\cite{xu2019multilevel} & C3D & 45.30 & 27.70 & 13.60 \\
  & TSP-PRL~\cite{wu2020tree} & C3D & 56.02 & 38.83 & - \\
  & TripNet~\cite{hahn2019tripping} & C3D & 48.42 & 32.19 & 13.93 \\
  & DRN~\cite{zeng2020dense} & C3D & - & 45.45 & 24.36 \\
  & CPNet~\cite{li2021proposal} & C3D & - & 40.56 & 21.63 \\
  & ABLR~\cite{yuan2019find} & C3D & 55.67 & 36.79 & - \\
  & DEBUG~\cite{lu2019debug} & C3D & 55.91 & 39.72 & - \\
  & ExCL~\cite{ghosh2019excl} & C3D & 63.00 & 43.60 & 24.10 \\
  & VSLNet~\cite{zhang2020span} & C3D & 63.16 & 43.22 & \textbf{26.16} \\
  \midrule
  \multicolumn{6}{c}{\textbf{Ours}} \\ \midrule
  \multirow{4}{*}{\begin{tabular}[c]{@{}c@{}} TVP-Based\\ 2D TVG\end{tabular}} &  Base w/o prompts & \multirow{4}{*}{ResNet} & 57.20  &  40.16 &  19.14 \\
  & Base + Visual Prompts &  &  60.12 & 43.39 & 23.71 \\
  & Base + Text Prompts &  &60.48 & 42.58 &  24.39\\
  & Base + Both Prompts &  & 60.71 & 43.44 &  25.03 \\
\midrule
\bottomrule[1pt]
\end{tabular}}
\label{tab:anet_result}
\end{table}
\vspace*{-4mm}

\noindent
\textbf{Effectiveness of TVP on Charades-STA.} 
The performance comparisons with SOTA methods on the Charades-STA dataset are summarized in \textbf{Tab.~\ref{tab:charades_result}}. 
Our proposed TVP framework can achieve competitive performance at all tIoU thresholds $m$ in the case of utilizing 2D visual features extracted by ResNet-50, and reach the highest score at $m=0.3$. Compared to the 2D TVG methods using VGG as the vision encoder, our proposed framework could achieve around 
$2.5\times$ and  $2.7\times$ performance gain 
at thresholds 
$0.5$ and $0.7$, respectively.
Furthermore, we can find that for our base model only one of visual prompts and text prompts can achieve up to $7.37\%$ and $9.60\%$ improvement at tIoU thresholds $m=0.3$ and $m=0.5$.
The combination of text and visual prompts can
not only achieves $7.55\%$ and $9.79\%$ improvements at tIoU thresholds $m=0.3$ and $m=0.5$, but also 
improve the performance by $8.14\%$  at $m=0.7$. This demonstrates the effectiveness and necessity of the joint text-visual prompting.

\begin{figure}[t]
    \centering
    \begin{subfigure}[t]{0.23\textwidth}
           \centering
           \includegraphics[width=\textwidth]{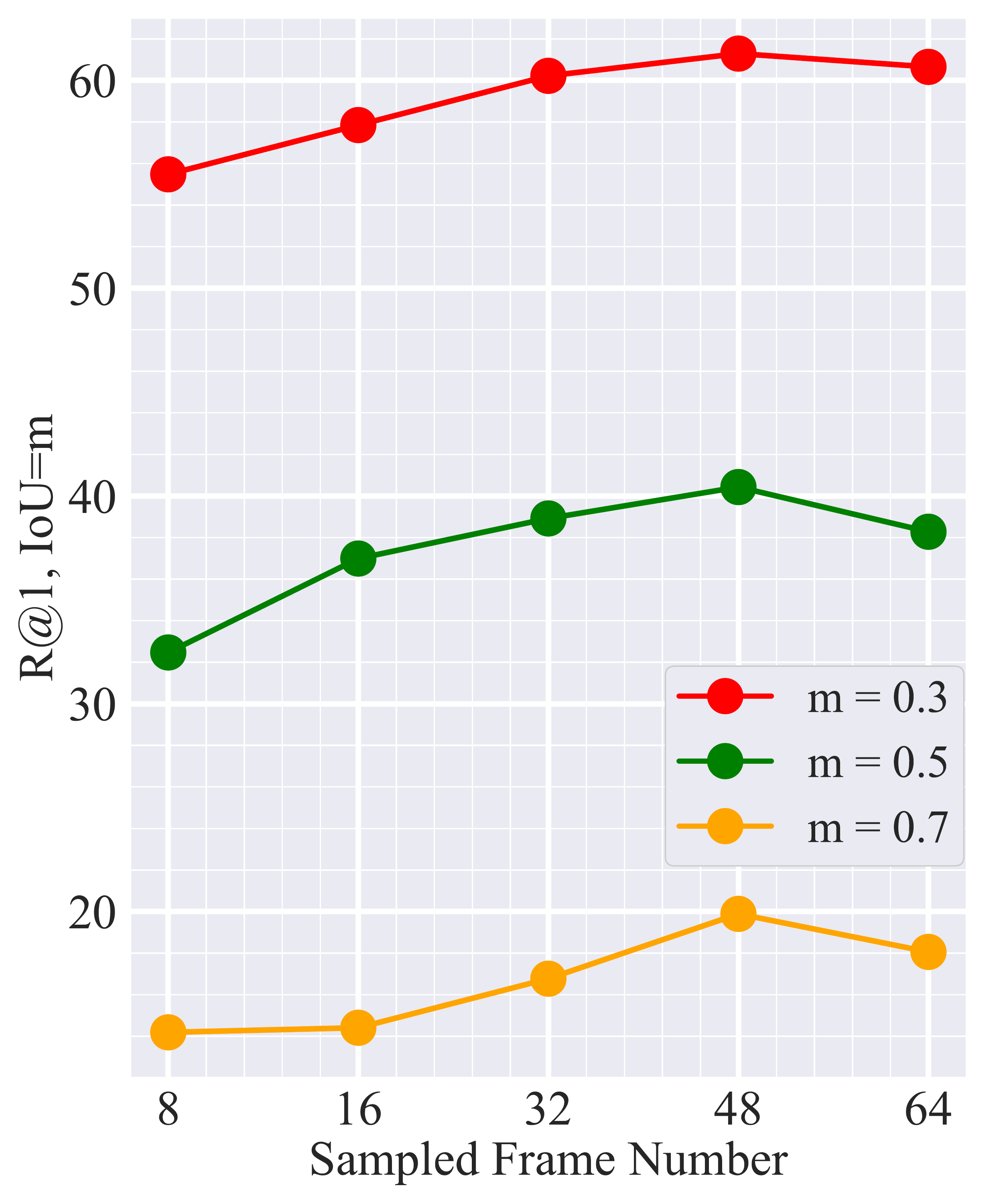}
            \caption{Charades-STA}
            \label{fig:4a}
    \end{subfigure}
    \begin{subfigure}[t]{0.23\textwidth}
            \centering
            \includegraphics[width=\textwidth]{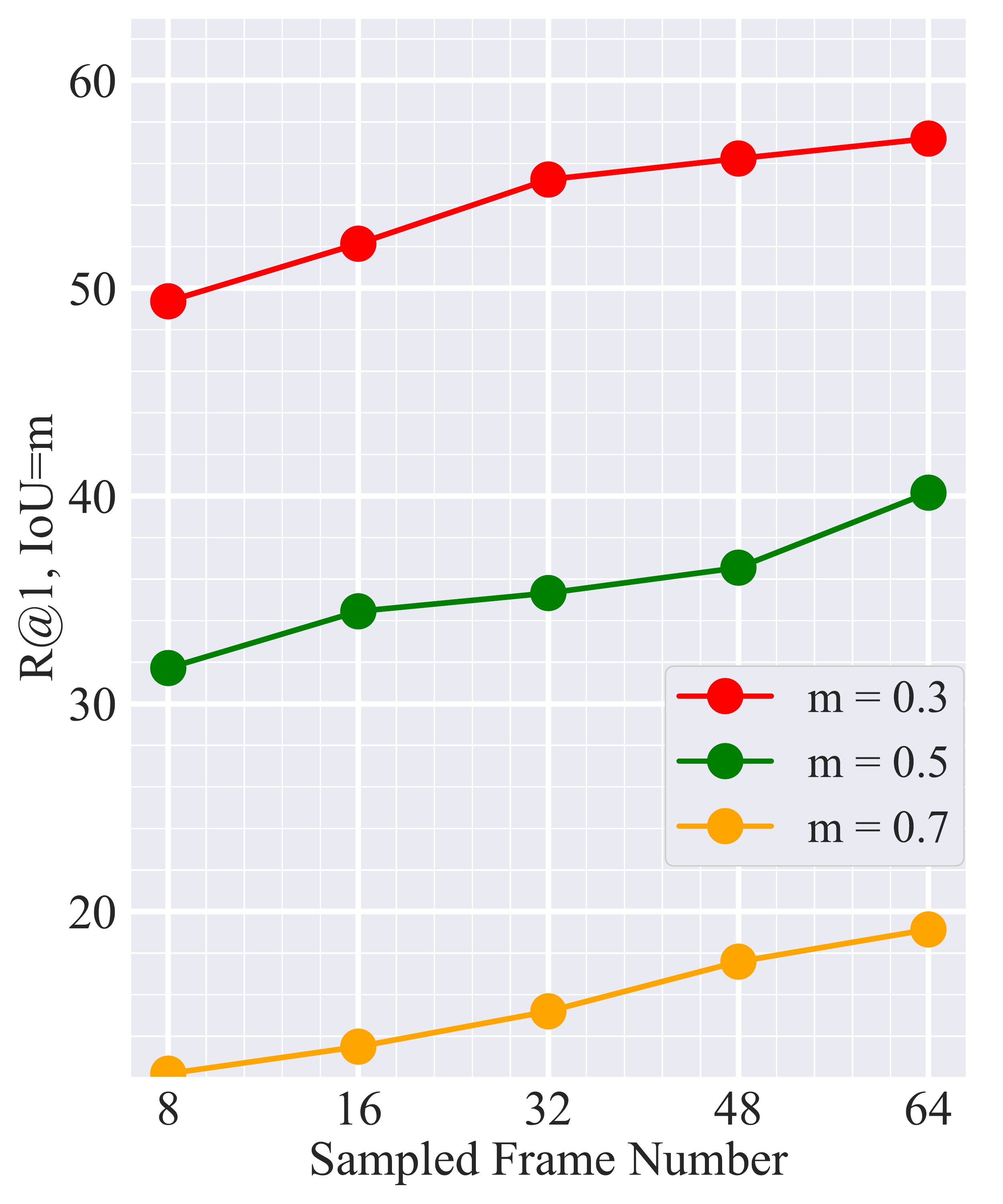}
            \caption{ActivityNet Captions}
            \label{fig:4b}
    \end{subfigure}
    \caption{Impact of sampled frame numbers.}
    \vspace*{-4mm}
\label{fig:frame_num}
\end{figure}

\noindent
\textbf{Effectiveness of TVP on ActivityNet Captions.} We focus on the performance comparisons with 3D TVG methods on ActivityNet since
there are no results of 2D TVG method reported on ActivityNet Captions.
The results of multiple methods on ActivityNet Captions datasets are reported in \textbf{Tab.~\ref{tab:anet_result}}. Even on this more challenging dataset, our proposed method still has achieved competitive performance compared to 3D TVG methods. Different from the performance of TVP on Charades-STA dataset, text prompts or visual prompts can achieve a significant performance boost on the base model over all IoU thresholds $m$ alone ($5.73\%$ at $m=0.3$, $8.04\%$ at $m=0.5$, $27.43\%$ at $m=0.7$ ) 
, and the text-visual prompt combination could further boost the performance ($6.14\%$ at $m=0.3$, $8.17\%$ at $m=0.5$, $30.77\%$ at $m=0.7$). It is worth noting that the performance gap over $m=0.7$ between 2D TVG methods and 3D TVG  methods is narrowed significantly.

\noindent
\textbf{In summary}, through the experimental results on Charades-STA and ActivityNet Captions datasets, we can find that our proposed TVP framework could achieve competitive performance overall tIoU thresholds on Charades-STA and ActivityNet Captions by improving the utility of sparse 2D visual features. Thanks to the lightweight 2D vision encoder, cotraining language encoder and vision encoder on large-scale image-text datasets can be performed, which benefits the base model to achieve good performance. Furthermore, the combination of text and visual prompts can achieve better results than any single kind of prompts on both datasets, which again proves the importance of crossmodal training.

\noindent
\textbf{Video frame sampling effect.} \textbf{Fig.~\ref{fig:frame_num}} demonstrates the performance of base model with different number $N_\mathrm{sam}$ of sampled video frames as visual inputs. For Charades dataset, the base model performance keeps increasing before $N_\mathrm{sam}$ reaches 48, but when it exceeds 48, performance starts to degrade. 
This is because 
frequent background changes harm the performance of object re-identification in videos, which are noisy for object motion analysis
\cite{gu2020appearance}.

For ActivityNet Caption dataset, base model performance continues to improve even when sampled frame number $N_\mathrm{sam}$ exceeds 48, due to the longer average video length in ActivityNet Captions dataset. Balancing the frame number and batch size for training, we choose $N_\mathrm{sam} = 64$ for ActivityNet Captions.

\begin{table}[t]
\caption{
The performance comparison of different visual prompt sizes on Charades-STA dataset. 
}
\vspace{-4mm}
\center
\resizebox{0.47\textwidth}{!}{
\begin{tabular}{c|ccc|c} 
\toprule
 \midrule
 Visual Prompt Size & & Acc(R@1, IoU=$m$) &  & Prompt + Frame\\
 & $m$=0.3 & $m$=0.5& $m$=0.7 \\
 \midrule
 0 & 61.29 & 40.43 & 19.89 & \includegraphics[width=2em, height=2em]{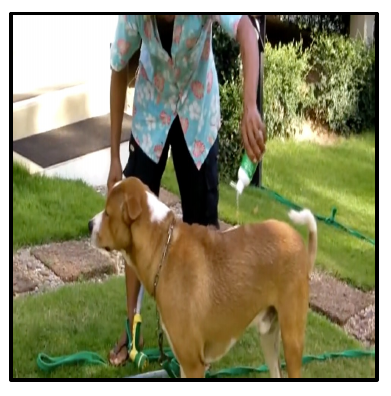} \\
 \midrule
 16 & 61.29 & 40.43 & 20.00 & \includegraphics[width=2em, height=2em]{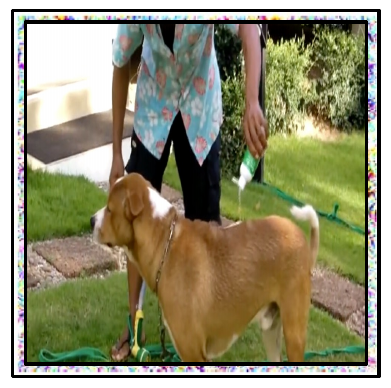} \\
 32 & 61.94 & 39.78 & 19.35 & \includegraphics[width=2em, height=2em]{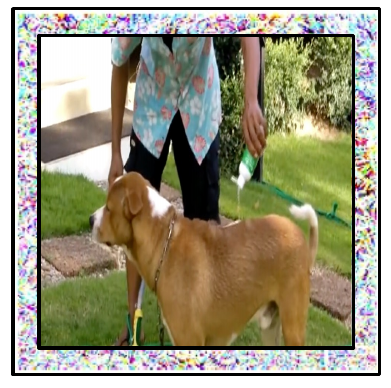} \\
 48 & 63.66& 42.37 & 20.00&     \includegraphics[width=2em, height=2em]{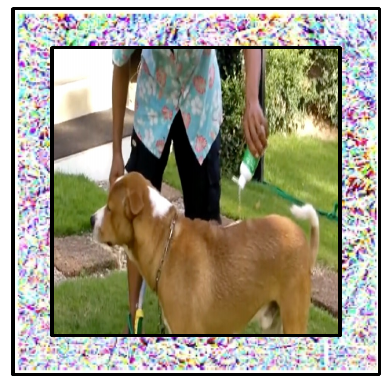} \\
 72 & 63.87 & 43.66 & 19.78 & \includegraphics[width=2em, height=2em]{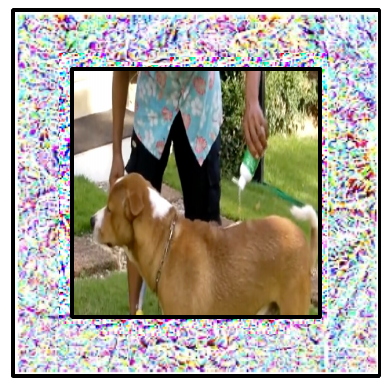} \\
 96 & \textbf{65.38} & \textbf{44.31} & \textbf{20.22} & \includegraphics[width=2em, height=2em]{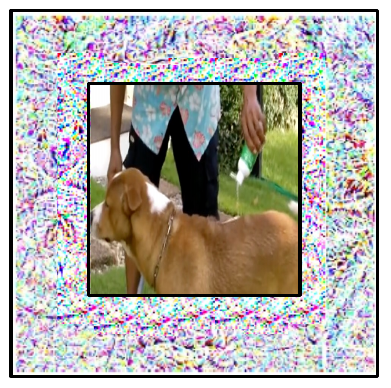} \\
 128 & 64.73& 43.66 & 19.78&  \includegraphics[width=2.2em, height=2.2em]{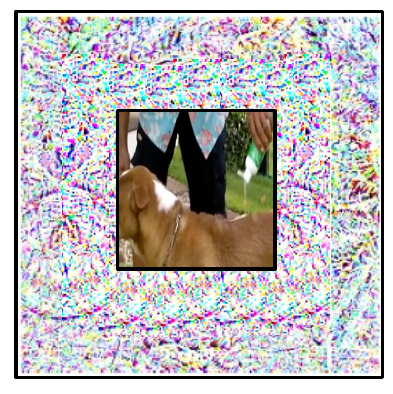}\\
 
\hline
\bottomrule
\end{tabular}}
\label{tab:pad_size}
\end{table}

\begin{table}[t]
\caption{
The performance comparison of different text prompt sizes on Charades-STA dataset.
}
\vspace{-2mm}
\centering
\resizebox{0.45\textwidth}{!}{
\begin{tabular}{c|ccc} 
\toprule
 \midrule
 Text Prompt Size & & Acc(R@1, IoU=$m$) &  \\
 & $m$=0.3 & $m$=0.5& $m$=0.7 \\
 \midrule
 0 & 57.20 & 40.16 & 19.14 \\
 \midrule
 5 & 65.38 & 41.94 & 20.43 \\
 10 & \textbf{65.81} & 43.44 & 20.65  \\
 15 & 65.59& 43.23 & 21.29 \\
 20 & 64.95 & \textbf{43.87} & \textbf{21.51}  \\
 25 & 63.66 & 42.80 & 20.65  \\
 30 & 64.46& 42.63 & 20.51\\
 
 \midrule
\bottomrule
\end{tabular}}
\vspace{-3mm}
\label{tab:text_size}
\end{table}

\noindent
\textbf{TVP performance vs. prompt size.} As shown in \textbf{Tab.~\ref{tab:pad_size}}, we can find that when visual prompts are small, they cannot bring changes to the base model, and when visual prompts are too large, the performance starts to decrease. This is because key information within video frames might be removed. However, the text prompts can bring significant performance boost even when the text prompt size is small as shown in \textbf{Tab.~\ref{tab:text_size}}, which is because the textual features has a smaller dimension compared to visual features, and also the text prompts are directly optimized in feature space during training.

\begin{table}[t]
\caption{
The performance comparison of different visual prompt operations (`\textit{remove}', `\textit{add}', `\textit{replace}') with fixed visual prompt size $p=96$
 on Charades-STA and ActivityNet Captions datasets.
}
\vspace{-4mm}
\center
\resizebox{0.47\textwidth}{!}{
\begin{tabular}{c|ccc|ccc} 
\toprule
 \midrule
 Operation & \multicolumn{3}{c|}{Charades-STA} & \multicolumn{3}{c}{ActivityNet Captions}\\
 \midrule
 & & R@1, IoU=$m$ & & &   R@1, IoU=$m$ &  \\
 & $m$=0.3 & $m$=0.5& $m$=0.7 & $m$=0.3 & $m$=0.5& $m$=0.7 \\
  \midrule
 Original & 61.29 & 40.43 & 19.89&  57.20& 40.16 & 19.14\\
  \midrule
 Remove & 61.29 & 40.43 & 20.0 & 57.20 & 40.16 & 19.14\\
 Add & 61.08 & 39.57 & 20.22 & 57.15 & 40.16 & 19.27 \\
 Replace & \textbf{65.38}& \textbf{44.31} & \textbf{20.22}& \textbf{60.12}& \textbf{43.39} &  \textbf{23.71}    \\
 
 \midrule
\bottomrule
\end{tabular}}
\label{tab:pad_operation}
\end{table}

\noindent
\textbf{TVP performance vs. visual prompt  operation.} 
Visual prompt is first proposed by \cite{bahng2022exploring}, where visual prompts are added to the image for transfer learning on classification tasks. In contrast, our proposed prompting framework is designed to compensate for the spatiotemporal information loss in 2D visual features. Due to the differences in the task, we try two different prompt operation strategies, `\textit{replace}' and `\textit{add}'. `\textit{add}' is to add the visual prompts to the pixel value of the video frame at the corresponding padding locations. `\textit{replace}' is to replace the pixel values of video frames with visual prompts at corresponding padding locations. `\textit{remove}' is in order to study the impact of removing the pixel values at the padding location. As shown in \textbf{Tab.~\ref{tab:pad_operation}}, `\textit{add}' or `\textit{remove}' prompt operations have limited effects on the base model. However, `\textit{replace}' does boost the base model performance.

\begin{figure}[t]
\centerline{\includegraphics[width=1.0\columnwidth]{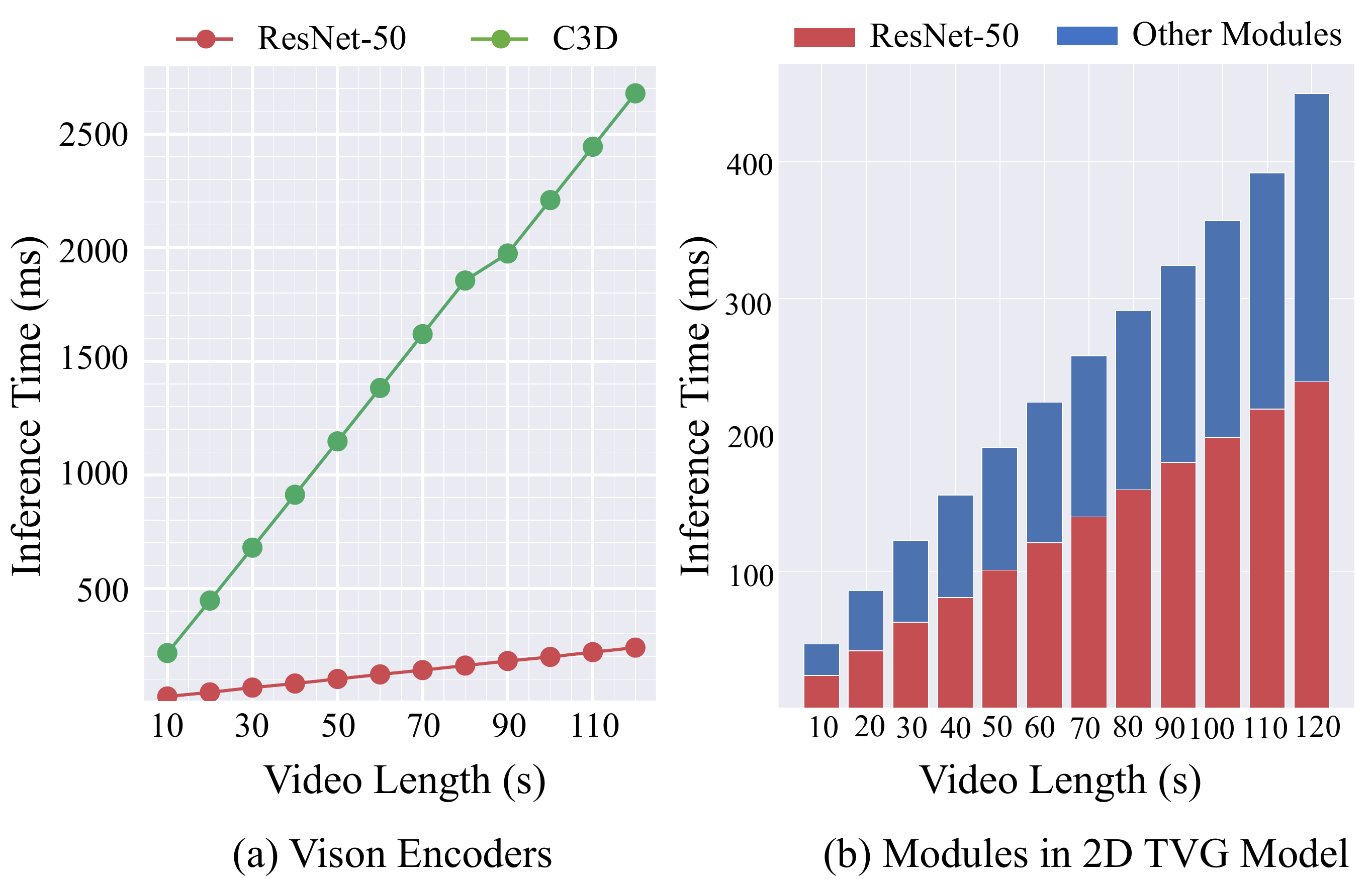}}
\caption{
Inference time comparison. (a) inference time comparison between 2D vision encoder (ResNet-50) and 3D vision encoder (C3D). (b)inference time comparison between the vision encoder and the other modules of the 2D TVG model, where the sampled frame number for our TVP framework is $1.2\times$ the length of the video in seconds. 
}
\label{fig: inference_time}
\end{figure}

\noindent
\textbf{TVP achieves inference efficiency.}  As shown in \textbf{Fig.~\ref{fig: inference_time}}, we can find that the inference time required for visual feature extraction accounts for more than half of the inference time of the whole model, while the inference time required for the 3D vision encoder is more than $5\times$ compared to the 2D vision encoder, and even more than the time required for the whole TVG model using 2D vision encoder, which fully demonstrates the feasibility of accelerating the overall inference speed by reducing the complexity of the vision encoder. Need to note that if there are multiple model weights for different sampled frame number settings and  model weights can be adopted adaptively for different lengths of videos, the inference speed for short videos should increase, and the prediction results for long videos will be further improved.

\begin{figure}[t]
  \vspace*{-2mm}
    \centering
    \begin{tabular}{cc}
    \hspace*{-3mm} \includegraphics[width=.225\textwidth,height=!]{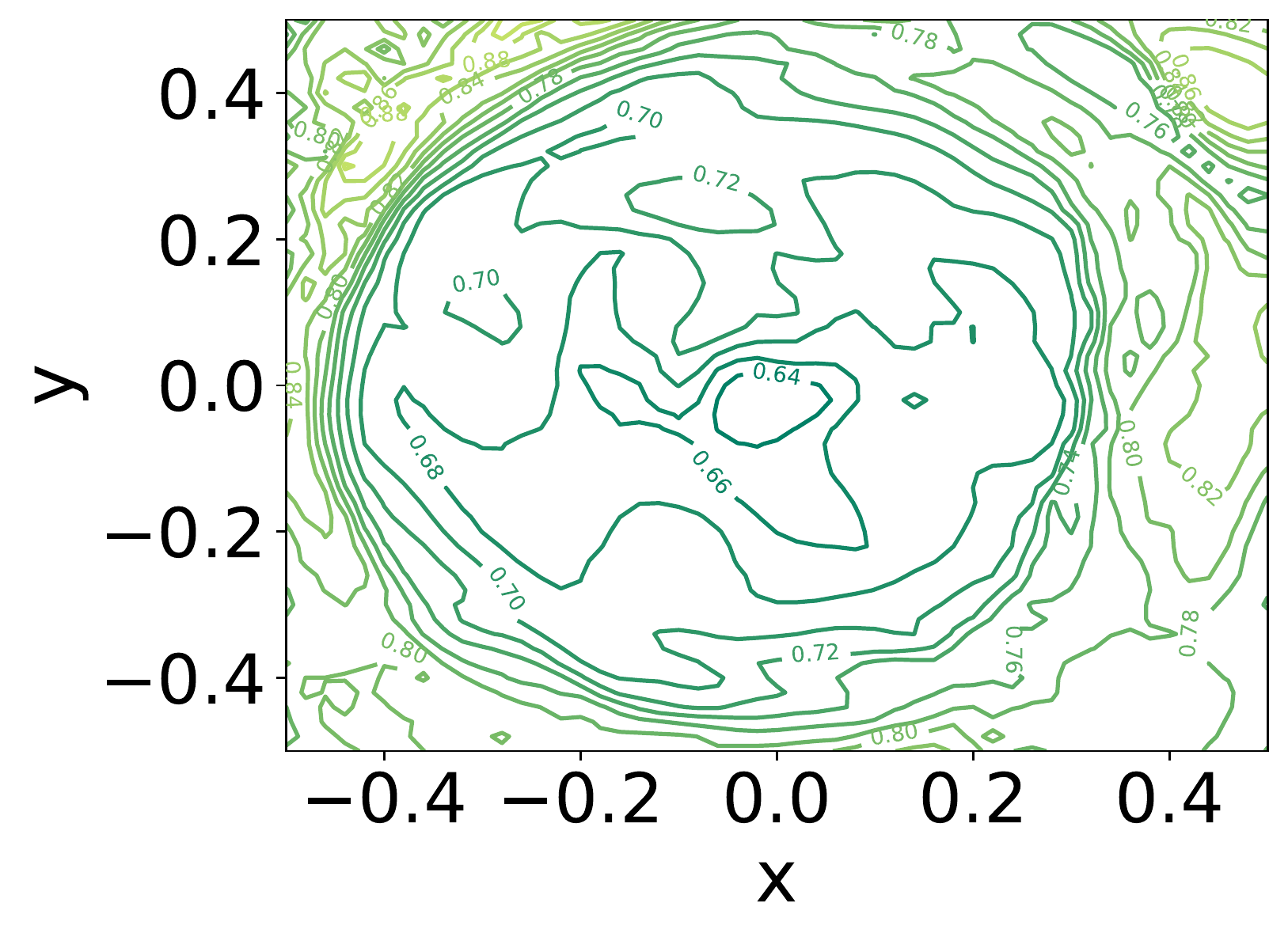}
     &
     \includegraphics[width=.225\textwidth,height=!]{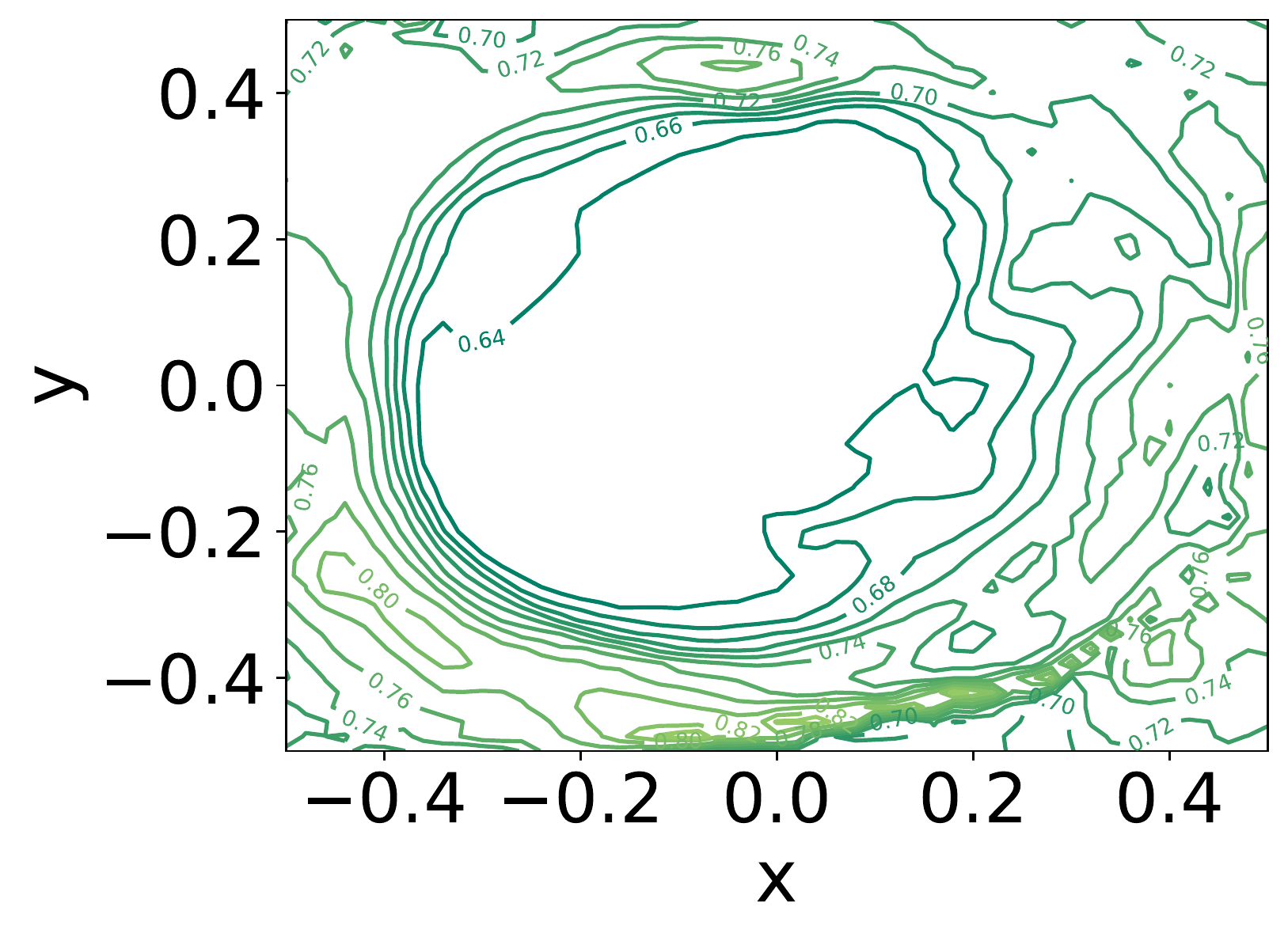}
    \end{tabular}
    \vspace*{-5mm}
    \caption{\footnotesize{Loss landscape visualization in 2D plane: Finetuning w/o prompts (left) and using prompts (right);  see \cite{li2018visualizing} for implementation.
    }}
    
     \label{fig:loss_landscape}
     \vspace*{-5mm}
\end{figure}

\noindent
\textbf{Ablation studies.} Through \textbf{Tab. \ref{tab:loss}}, we can find that the addition of either distance loss $\mathcal{L}_\mathrm{dis}$ or duration loss  $\mathcal{L}_\mathrm{dur}$ will result in a performance increase, but the combination of the two will result in a significant performance increase ($11.34\%$ at $m=0.3$, $35.26\%$ at $m=0.5$, $68.27\%$ at $m=0.7$, ), especially over tIoU thresholds $m=0.5$ and $m=0.7$. This demonstrates that distance loss $\mathcal{L}_\mathrm{dis}$ and duration loss $\mathcal{L}_\mathrm{dur}$ could provide more precise training guides compared to only using temporal IoU loss $\mathcal{L}_\mathrm{tIoU}$. Furthermore, we posit that prompting may encode additional spatial-temporal supervision to help the model trainer to escape from bad local optima as shown in Fig.~\ref{fig:loss_landscape}, where fine-tuning  w/ prompts yields a flatter loss landscape than the one w/o prompts.

\begin{table}[t]
\caption{
The performance comparison of different loss designs on Charades-STA dataset.
}
\vspace{-4mm}
\center
\resizebox{0.42\textwidth}{!}{
\begin{tabular}{c|ccc} 
\toprule
 \midrule
 Loss Function Selection & & R@1, IoU=$m$ &  \\
 & $m$=0.3 & $m$=0.5& $m$=0.7 \\
 \midrule
 $\mathcal{L_\mathrm{tIoU}}$ & 55.05  & 29.89 & 11.82 \\
 \midrule
 $\mathcal{L_\mathrm{tIoU}} + \mathcal{L_\mathrm{dis}}$ & 60.64 & 31.18 & 16.77 \\
 $\mathcal{L_\mathrm{tIoU}} + \mathcal{L_\mathrm{dur}}$ & 59.78 & 30.97 & 16.34  \\
 $\mathcal{L_\mathrm{tIoU}} + \mathcal{L_\mathrm{dis}} + \mathcal{L_\mathrm{dur}}$ & \textbf{61.29} & \textbf{40.43} & \textbf{19.89}  \\
 
 \midrule
\bottomrule
\end{tabular}}
\label{tab:loss}
\end{table}

\section{Conclusion}
\label{sec:conclusion}
In this paper,
we propose text-visual prompting to boost the performance of 2D TVG methods by compensating for the lack of spatiotemporal information in 2D visual features. In   contrast to 3D TVG methods, TVP allows us to effectively co-train vision encoder and language encoder in a
2D TVG model and improves the performance of cross-
modal feature fusion using only low-complexity sparse 2D
visual features. 
The effectiveness of our proposed TVP~(text-visual prompting) framework has been
demonstrated on two standard datasets, Charades-STA and ActivityNet. Our models outperform all 2D models significantly, and also achieve comparable performance to 3D models. What is more, we achieve over $5 \times$ inference speedup over TVG methods of using 3D visual features.

{\small
\bibliographystyle{ieee_fullname}
\bibliography{egbib}
}

\end{document}